\documentclass[runningheads]{llncs}

\usepackage{eccv}

\usepackage{eccvabbrv}

\usepackage{graphicx}
\usepackage{booktabs}

\usepackage[accsupp]{axessibility}  

\usepackage{hyperref}

\usepackage{orcidlink}

\usepackage{amsmath,amsfonts,bm}

\def\eqref#1{equation~\ref{#1}}

\def\1{\bm{1}}

\DeclareMathAlphabet{\mathsfit}{\encodingdefault}{\sfdefault}{m}{sl}
\SetMathAlphabet{\mathsfit}{bold}{\encodingdefault}{\sfdefault}{bx}{n}

\definecolor{mycyan}{RGB}{212, 239, 251}
\usepackage{colortbl}
\usepackage{xcolor}
\definecolor{mygray}{gray}{.9}
\definecolor{goldenrod}{RGB}{245,245,220}
\newlength\savewidth
\newcolumntype{a}{>{\columncolor{mygray}}c}
\usepackage{fontawesome5}
\usepackage{booktabs}

\usepackage{fontawesome5}
\usepackage{lipsum}
\usepackage{comment}
\usepackage{multirow}
\usepackage{wrapfig}
\usepackage{algorithm}
\usepackage{algorithmic}
\usepackage{xfrac}  

\usepackage{graphicx}
\usepackage{color}
\usepackage{makecell}
\definecolor{darkgreen}{rgb}{0,0.7,0}

\definecolor{mygraytext}{gray}{.5}
\newcommand{\graytext}[1]{{\color{mygraytext}{#1}}}

\begin{document}

\title{Active Generation for Image Classification} 


\author{Tao Huang\inst{1}\thanks{The authors contributed equally. $^{\dag}$ Corresponding author.} \and
Jiaqi Liu\inst{1}$^*$ \and Shan You\inst{2}$^{\dag}$ \and
Chang Xu\inst{1}}

\authorrunning{T.~Huang et al.}

\institute{School of Computer Science, Faculty of Engineering, The University of Sydney\\
\email{\{thua7590,jliu6979\}@uni.sydney.edu.au, c.xu@sydney.edu.au}
\and
SenseTime Research\\
\email{youshan@sensetime.com}
}

\maketitle

\begin{abstract}
  Recently, the growing capabilities of deep generative models have underscored their potential in enhancing image classification accuracy. However, existing methods often demand the generation of a disproportionately large number of images compared to the original dataset, while having only marginal improvements in accuracy. This computationally expensive and time-consuming process hampers the practicality of such approaches. In this paper, we propose to address the efficiency of image generation by focusing on the specific needs and characteristics of the model. With a central tenet of active learning, our method, named ActGen, takes a training-aware approach to image generation. It aims to create images akin to the challenging or misclassified samples encountered by the current model and incorporates these generated images into the training set to augment model performance. ActGen introduces an attentive image guidance technique, using real images as guides during the denoising process of a diffusion model. The model's attention on class prompt is leveraged to ensure the preservation of similar foreground object while diversifying the background. Furthermore, we introduce a gradient-based generation guidance method, which employs two losses to generate more challenging samples and prevent the generated images from being too similar to previously generated ones. Experimental results on the CIFAR and ImageNet datasets demonstrate that our method achieves better performance with a significantly reduced number of generated images. Code is available at \url{https://github.com/hunto/ActGen}.
\end{abstract}

\begin{figure}[tb]
  \centering
  \begin{subfigure}{0.4\linewidth}
    \includegraphics[width=\linewidth]{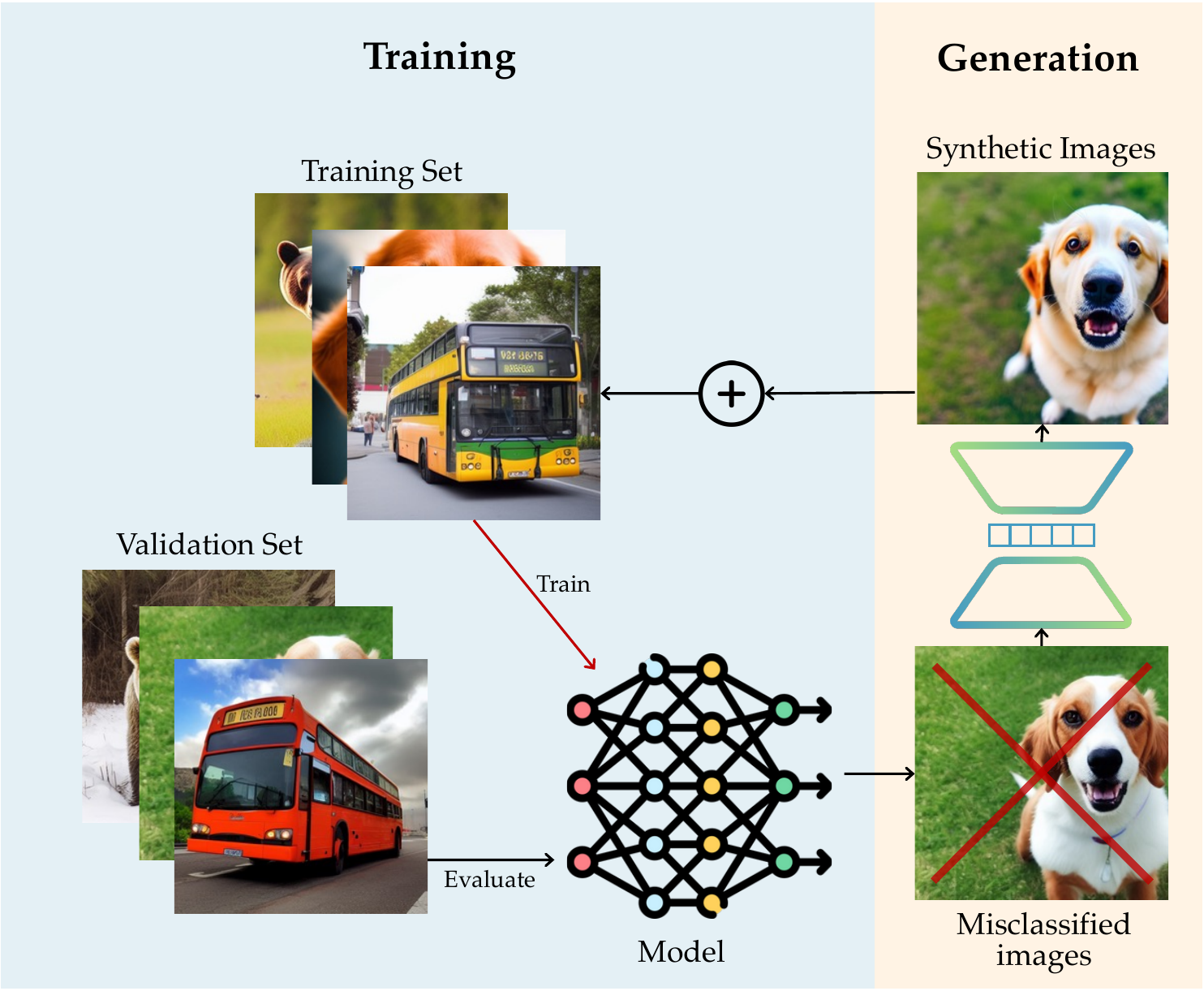}
    \caption{}
    \label{fig:framework}
  \end{subfigure}
  \hfill
  \begin{subfigure}{0.55\linewidth}
    \includegraphics[width=\linewidth]{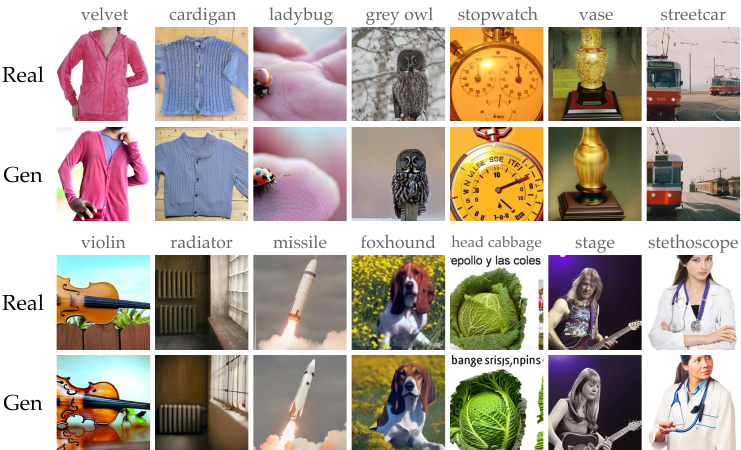}
    \caption{}
    \label{fig:hard_samples}
  \end{subfigure}
  \vspace{-2mm}
  \caption{(a) Illustration of the process of active generation. Misclassified images are utilized as guides for generating hard samples, which are then incorporated into the training set. (b) Examples of misclassified images in ImageNet dataset. Our ActGen can augment the hard samples to similar ones.}
  \vspace{-4mm}
\end{figure}

\section{Introduction}
\label{sec:intro}

The rapid advancements in deep learning, propelled by extensive training data and automated feature engineering, have brought significant breakthroughs in computer vision tasks, including image recognition~\cite{simonyan2014very,he2016deep,dosovitskiy2020image}, object detection~\cite{ren2015faster,lin2017focal, carion2020end}, and semantic segmentation~\cite{long2015fully,zhao2017pyramid,chen2018encoder}. Despite these achievements, the manual collection of large-scale labeled datasets remains a costly and time-consuming endeavor. Furthermore, concerns related to data privacy and usage rights have introduced additional hurdles in the acquisition of such datasets.

Recently, deep generative models \cite{ho2020denoising,Rombach_2022_CVPR,saharia2022photorealistic} have made remarkable strides, driven by increased model capacity and access to larger datasets. These advancements have enabled the generation of high-fidelity images that correspond to specific conditions, such as text descriptions or predefined classes. Consequently, researchers have explored the use of synthetic data generated by these models in image classification and few-shot image classification tasks \cite{he2023is,azizi2023synthetic,zhou2023training}. However, despite their potential, these approaches often yield only marginal improvements over baseline models trained solely on real data. The primary drawback lies in their voracious appetite for synthetic images, resulting in substantial computational costs and energy consumption\footnote{Stable diffusion V2~\cite{Rombach_2022_CVPR} costs about $3$ seconds and $16$ TMACs to generate a $512\times512$ image on a V100 GPU.}. For instance, \cite{azizi2023synthetic} reported a mere 1.78\% increase in accuracy in ImageNet classification, achieved by augmenting the real dataset with an equivalent number of 1.2 million synthetic images. This stark trade-off between computational resources and performance improvement raises a critical challenge in the application of synthetic data for image classification.

In this paper, we assert that the inefficiency of existing methods arises from the unrestricted generation of images, resulting in a significant proportion of redundant images when compared to the target dataset. Therefore, we advocate a shift towards a more precise approach — prioritizing the generation of images specifically demanded by the model to enhance its performance on the target dataset. Our approach incorporates active learning \cite{ren2021survey} as a central tenet. By partitioning a validation set from the training data, we utilize these validation samples to continuously assess the model's performance throughout training. When the model misclassifies validation images, it acts as a signal, pinpointing areas where the model lacks proficiency. To bolster validation accuracy, we strategically augment the misclassified images, enabling the model to specifically address these challenging cases. This active learning strategy ensures that the model focuses on refining its performance in areas crucial for optimal results on the target dataset.

To augment misclassified images while preserving their inherent characteristics and infusing diverse scenes, we introduce an innovative approach dubbed attentive image guidance. This method leverages real images to guide the diffusion generation process. At each timestep within the DDPM sampler~\cite{ho2020denoising}, it interpolates the generated latent feature with the latent feature of a real image, producing a novel latent feature that encapsulates the desired characteristics. Furthermore, we harness the attentions within the cross-attention layer to precisely locate the foreground instance. These attentions are then employed as masks, restricting interpolations to the foreground areas. This targeted approach not only maintains the integrity of the instance but also facilitates the generation of diverse background scenes.

In addition, to enhance the diversity of generated images and exert more complex control over the generation process, we introduce a gradient-based generation guidance mechanism. Unlike direct feature interpolation, this method enables us to achieve more nuanced control over the generation process. The mechanism involves backpropagating losses computed on the generated latents to the input text embedding, which serves as a critical conditional signal to steer text-to-image diffusion models. By updating the embedding. We apply two key types of losses to refine the text embedding: (1) Contrastive loss: This loss quantifies the distances between the current latent feature and the latent features of previously generated images. It acts as a regularizer, preventing the current image from closely resembling previous ones, thereby reducing redundancy. (2) Classification loss: The loss seeks to maximize the prediction loss of the current classification model. This approach challenges the diffusion model to generate images that are more difficult to classify, enhancing the overall quality of generated content. 

In summary, our contributions can be categorized into three key areas:
\begin{enumerate}
    \item We introduce an inventive approach to generate images from misclassified validation data during training, coupled with an attentive image guidance mechanism. This strategy enhances the practicality of generated images by aligning them with the model's evolving needs throughout the training process.
    \item To further diversify synthetic images and gain finer control over the generation process, we present a gradient-based guidance mechanism. This mechanism refines the text embedding through two essential losses for generating a wider variety of images and elevating the classification difficulty.
    \item We conduct extensive experiments across various image classification settings to showcase the effectiveness of ActGen. For instance, on ImageNet classification, compared to previous work \cite{azizi2023synthetic}, ActGen utilizes only 10\% (0.13 million) of the synthetic images,  while achieving a notable 2.26\% accuracy improvement on ResNet-50 over the baseline. 
\end{enumerate}

\section{Related Work}

\subsection{Diffusion Models}

The diffusion model, originally introduced by \cite{sohl2015deep}, is based on principles derived from non-equilibrium thermodynamics. This model was initially proposed to establish the feasibility of sampling from a complex probabilistic distribution, thereby establishing the fundamental framework. The model was subsequently enhanced by DDPM~\cite{ho2020denoising}, which introduced variational inference for training and incorporated a parametrized neural network as a denoiser in the backward diffusion process. The log-likelihood score of the DDPM may not adequately capture the distribution of real data. As a solution, DDIM~\cite{nichol2021improved} was introduced to incorporate a learned variance for diffusion sampling.

State-of-the-art generative diffusion models, such as stable diffusion \cite{Rombach_2022_CVPR}, GLIDE \cite{nichol2021glide}, DALLE-3 \cite{shi2020improving}, and Imagen \cite{saharia2022photorealistic}, have demonstrated remarkable generative capabilities by producing text-conditioned high-fidelity photo-realistic synthetic samples. The controllability by ensuring that the generated image content aligns with the input prompt can also be regarded as a constraint, compelling the generated content to be relevant to the provided class label. This allows generating images that are related to specific classes for image classification tasks. One significant concern arises from the fact that models are trained on diverse datasets, leading to variations in their generative capabilities across different subjects. 

\subsection{Training with Synthetic Images}

The utilization of synthetic data generated by generative models, which have the capability to produce high-fidelity photo-realistic images, has facilitated a few studies aiming to enhance classification accuracy. Before diffusion models gained prominence, Generative Adversarial Networks (GANs) \cite{goodfellow2014generative, guo2020positive}, served as the primary generative framework for creating synthetic images for classification tasks. \cite{zhang2021datasetgan} leveraged the latent space of GANs, specifically StyleGAN \cite{karras2019style}, to produce diverse images with corresponding labels. However, \cite{besnier2020dataset} pointed out that earlier efforts, such as those by \cite{ravuri2019seeing} using BigGAN \cite{brock2018large} — a model capable of generating images across all 1000 ImageNet classes—observed a significant decline in classifier performance when trained on these synthetic images. To mitigate the performance drop in classifiers trained with synthetic data, \cite{besnier2020dataset} proposed three strategies: optimizing latent codes, employing continuous sampling, and adapting at test time. These methods are designed to improve the diversity and quality of synthetic data, thereby enhancing classifier accuracy.

The superior performance of generative diffusion models \cite{dhariwal2021diffusion} has shifted the focus of using synthesized images for classification towards manipulating three key perspectives of the diffusion process: image, text, and model. By steering the diffusion process with real images using real guidance, \cite{he2023is} achieves a few-shot performance that is state-of-the-art across multiple datasets. \cite{shipard2023diversity} proposes a bag of tricks for addressing model-agnostic zero-shot problems by manipulating the input prompt and guidance scale and demonstrates how diversity impacts the efficacy of synthetic data. By learning text embedding for each image and perturbing them with random noise or via linear interpolation with other embeddings, \cite{zhou2023training} is capable of generating diverse alternation while preserving the overall integrity of the original image. At the model level, \cite{azizi2023synthetic} generates images utilizing an ImageNet fine-tuned diffusion model, which is then applied to ImageNet classification tasks. However, these methods employ a separate two-step strategy that first generate images then train the model, and require a substantial quantity of generated images in order to attain improvements on performance. In contrast, our study proposes a unified active generation framework and attains comparable or potentially superior performance while utilizing a significantly fewer number of samples.

\section{Preliminaries}

\subsection{Denoising Diffusion Probabilistic Models}

Diffusion models represent a class of probabilistic generative models that gradually introduce noise to sample data and subsequently learn to reverse this process by predicting and removing the noise. Formally, starting with the sample data $\bm{x}_0\in\mathbb{R}^{C\times H\times W}$, the forward noise process iteratively adds Gaussian noise to it:
\begin{equation} \label{eq:diff_process}
    q(\bm{x}_t|\bm{x}_0) := \mathcal{N}(\bm{x}_t|\sqrt{\bar{\alpha}_t}\bm{x}_0,(1 - \bar{\alpha}_t)\bm{I}),
\end{equation}
where $\bm{x}_t$ represents the transformed noisy data at timestep $t \in \{0,1,...,T\}$, and 
$\bar{\alpha}_t := \Pi_{s=0}^t\alpha_s = \Pi_{s=0}^t(1-\beta_s)$  
$\bar{\alpha}_t$ is a predetermined notation for the direct sampling of $\bm{x}_t$ at arbitrary timestep with a noise variance schedule $\beta$~\cite{ho2020denoising}. Therefore, we can express $\bm{x}_t$ as a linear combination of $\bm{x}_0$ and noise variable $\bm{\epsilon}_t$:
\begin{equation}
    \bm{x}_t = \sqrt{\bar{\alpha}_t}\bm{x}_0 + \sqrt{1 - \bar\alpha_t}\bm{\epsilon}_t,
\end{equation}

where $\bm{\epsilon}_t\in\mathcal{N}(\bm{0}, \bm{I})$. During training, a neural network is trained to predict the noise $\bm{\epsilon}_\theta(\bm{x}_t, t)$ in $\bm{x}_t$ \textit{w.r.t.} $\bm{x}_0$ by minimizing the L2 squared loss between $\bm{\epsilon}_\theta(\bm{x}_t, t)$ and $\bm{\epsilon}_t$.

During inference, with the initial noise $\bm{x}_t$, the data sample $\bm{x}_0$ is reconstructed with an iterative denoising process using the trained network:
\begin{equation} \label{eq:rev_diff_process}
    p_\theta(\bm{x}_{t-1}|\bm{x}_t) := \mathcal{N}(\bm{x}_{t-1}; \bm{\epsilon}_\theta(\bm{x}_t, t), \sigma_t^2\bm{I}),
\end{equation}
where $\sigma_t^2$ denotes the transition variance in DDPM~\cite{ho2020denoising}.

\begin{figure*}[t]
    \centering
    \includegraphics[width=\linewidth]{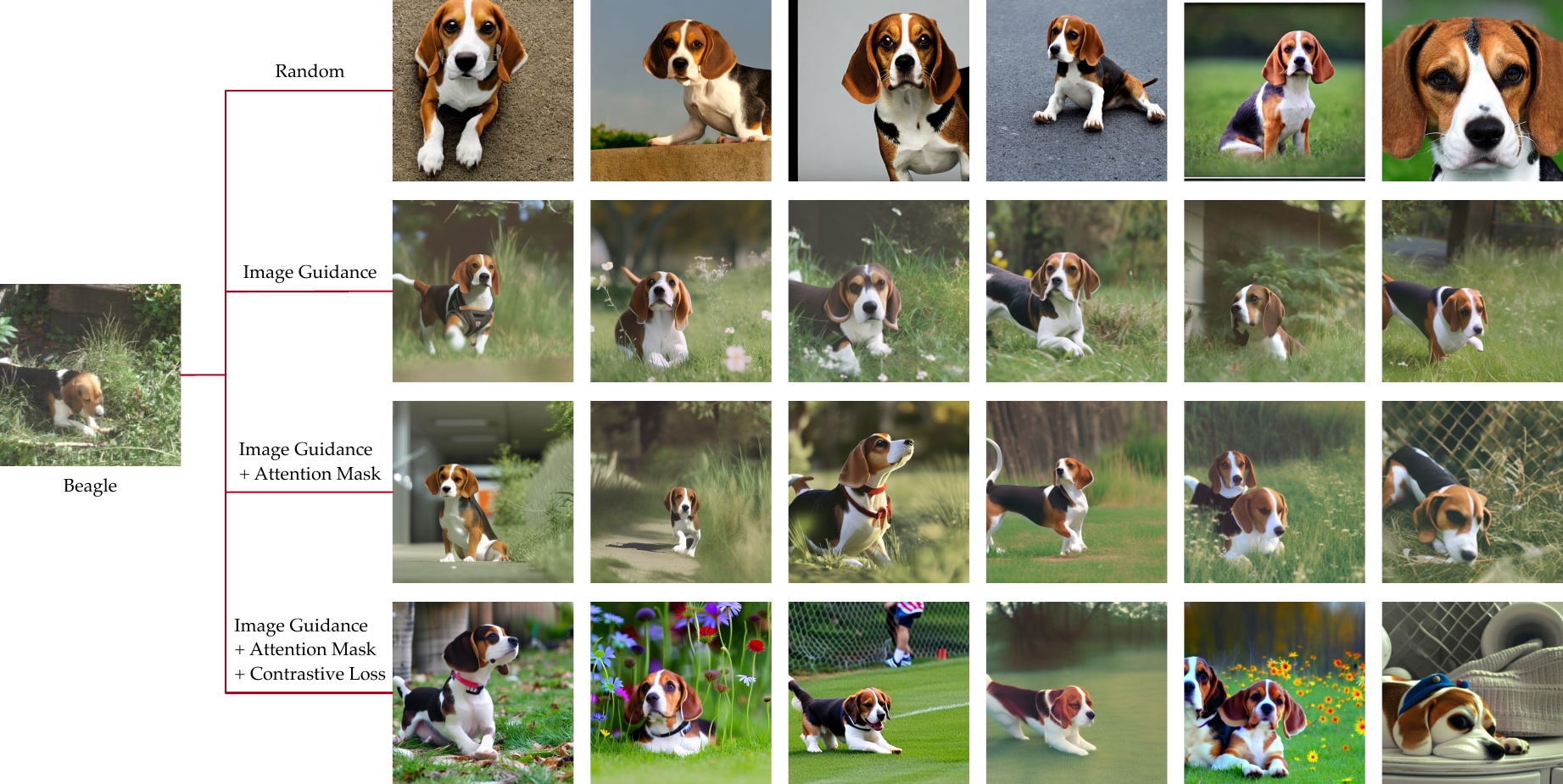}
    \caption{Visualizations of different generation guidance methods. Random: Random generalization on SD with a fixed prompt \texttt{a photo of beagle}. The last three rows are our methods with different proposed guidance mechanisms.}
    \label{fig:vis_cmp}
    \vspace{-4mm}
\end{figure*}

\subsection{Text-Conditioned Guidance}

Text-to-image diffusion models are generative models designed to create realistic images from textual descriptions. These models employ diffusion processes to iteratively generate images based on the semantic information provided in the input text. The objective is to capture the essence of the textual description and translate it into visually coherent images.

Text-to-image diffusion models incorporate additional text conditional variables $\bm{c}$ in the noise prediction model to predict the conditional noise $\bm{\epsilon}_\theta(\bm{x}_t, \bm{c}, t)$ and guide the generation. A classifier-free guidance technique \cite{ho2021classifier} is typically adopted, enabling the utilization of text-conditioned guidance during training without the need for a classifier. The denoising model is trained to handle both conditioned input, where a text prompt is provided, and unconditioned input, where the prompt is replaced with $\emptyset$. This allows for the representation of the guidance direction as the translation from the conditioned input $\bm{\epsilon}_\theta(\bm{x}_t, \bm{c}, t)$ to the unconditioned input $\bm{\epsilon}_\theta(\bm{x}_t, \emptyset, t)$. The guidance is performed with a guidance scale $s$ by
\begin{equation}
    \bm{\hat{\epsilon}}_\theta(\bm{x}_t, \bm{c}, t) = \bm{\epsilon}_\theta(\bm{x}_t, \emptyset, t) + s \left(\bm{\epsilon}_\theta(\bm{x}_t, \bm{c}, t) - \bm{\epsilon}_\theta(\bm{x}_t, \emptyset, t)\right).
\end{equation}

In this paper, we directly use the public text-to-image diffusion models~\cite{Rombach_2022_CVPR,nichol2021glide} trained on large-scale text-image pairs as our generative models. Based on the fixed text condition \texttt{a photo of <class>}, we further propose additional guidance methods to control the generation in inference process.

\section{Method}

\subsection{Active Generation of Hard Samples}

In the context of enhancing image classification with synthetic samples, existing approaches often adopt a two-stage strategy. This involves the initial generation of samples, followed by the integration of these generated samples into the model training process. Consequently, these works primarily focus on strategies to generate diverse and sufficient samples for improved training: \cite{azizi2023synthetic} proposes to finetune the diffusion model with target dataset to reduce the domain gap between real and synthetic images, \cite{zhou2023training} proposes multiple methods to generate diverse alternations of the original image, \textit{e.t.c}. However, it's important to note that these model-agnostic generation methods still face challenges, \textit{i.e.}, they lack the ability to discern which samples are genuinely beneficial to the model. Consequently, they often necessitate a substantial number of generated samples to achieve meaningful improvements.

In contrast, our approach is meticulously designed to maximize performance gains while utilizing as few generated samples as possible. This objective aligns with the principles of active learning, a paradigm that seeks to identify the most beneficial samples from a dataset for constructing the training dataset. The determination of which samples are truly helpful to the model is well-explored in both active learning and curriculum learning~\cite{bengio2009curriculum}. The latter, inspired by human education, stands out as a prominent technique for expediting the training process by gradually increasing the difficulty level of training samples.

In active learning and curriculum learning, the selection of useful samples is pivotal. Interestingly, it is recognized that the model converges more rapidly when trained on batches of challenging samples compared to randomly selected batches \cite{loshchilov2015online,schaul2015prioritized,song2020carpe}. Therefore, in our quest to generate only a fraction of images yet with decent benefits, it is also natural to consider generating those challenging samples for the target model.

\begin{figure*}[t]
    \centering
    \includegraphics[width=\linewidth]{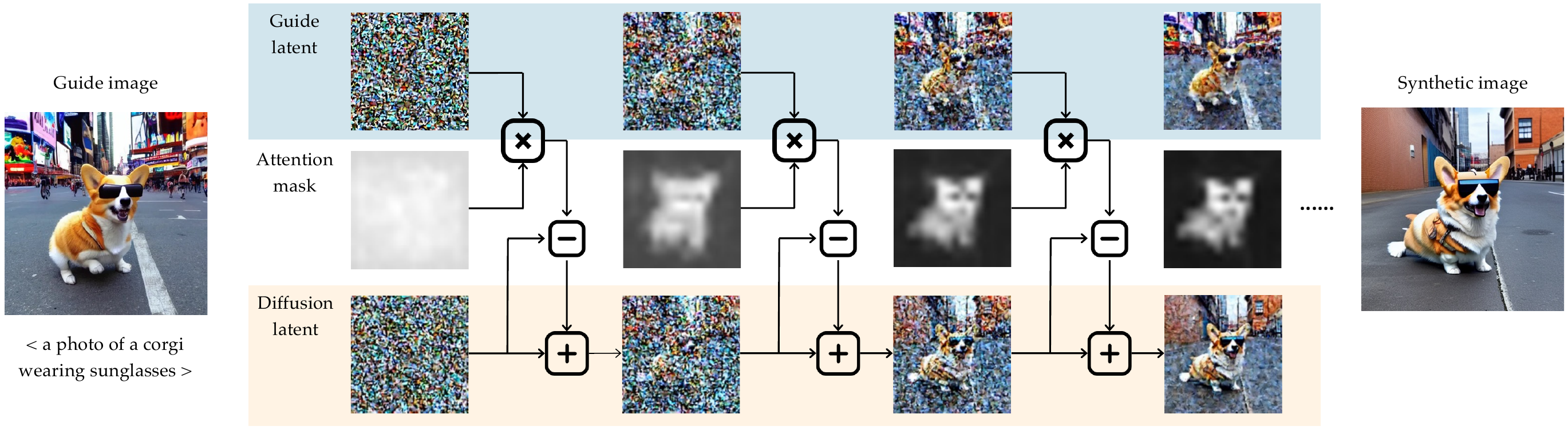}
    \caption{Illustration of attentive image guidance. It iteratively perturbs the diffusion latent with an attention mask at every timestep.}
    \label{fig:image_guidance}
    \vspace{-4mm}
\end{figure*}

We identify challenging samples by evaluating the model on a dedicated validation dataset, which is partitioned from the training set. The instances misclassified by the model serve as prototypes for hard samples. As visualized in Figure~\ref{fig:hard_samples}, the misclassified images exhibit unusual characteristics such as incomplete objects, extraordinary poses, and uncommon patterns within their categories. These rare samples in the training dataset contribute to the model's misclassification. Consequently, our objective is to obtain a model that generalizes well to these rare and challenging samples. To achieve this, we encourage augmenting these challenging images using generative models, effectively expanding the dataset to include variations of these prototypes. By doing so, the model is guided to specialize in handling these intricate instances, ultimately enhancing its accuracy. As illustrated in Figure \ref{fig:framework}, the training of the classification model commences with the real image dataset. After each epoch, we employ the validation set to identify misclassified samples, which are then utilized as guides for generating images in the diffusion models. Subsequently, these generated images are incorporated into the training set for further refinement through subsequent training epochs. In the subsequent sections, we will present a comprehensive introduction to our proposed mechanisms for guiding the generation of images.

\begin{figure*}[t]
    \centering
    \includegraphics[width=\linewidth]{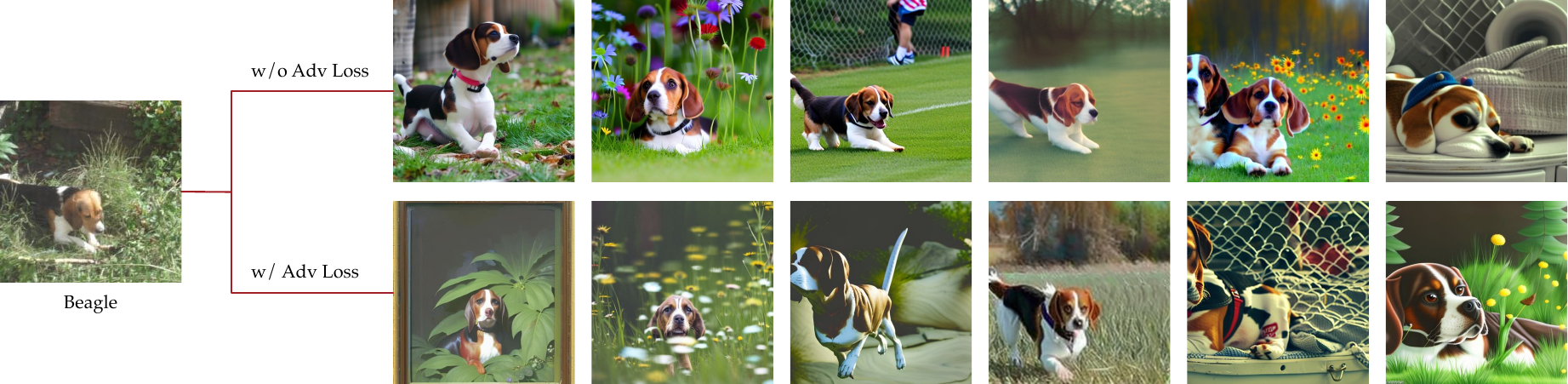}
    \caption{Visualizations of normal synthetic images and adversarial samples. Our adversarial samples have higher classification difficulty by blurring, occluding, or changing the contrast and style of the image.}
    \label{fig:vis_adv}
    \vspace{-4mm}
\end{figure*}

\subsection{Attentive Image Guidance}

To augment the provided hard image samples, our objective is to generate new images with similarities to the existing ones. We accomplish this by introducing real hard images into the generative denoising process of diffusion models.

At each denoising step of DDPM sampler \cite{ho2020denoising}, given the latent variable $\bm{x}_t$ and predicted noise $\bm{\hat{\epsilon}}_\theta(\bm{x}_t, \bm{c}, t)$, it samples $\bm{x}_{t-1}$ by
\begin{equation}
    \bm{x}_{t-1} = \frac{1}{\sqrt{\alpha_t}} \left(\bm{x}_t - \frac{1-\alpha_t}{\sqrt{1-\bar\alpha_t}}\bm{\hat{\epsilon}}_\theta(\bm{x}_t, \bm{c}, t)\right) + \sigma_t \bm{\mathrm{z}},
\end{equation}
where $\alpha_t$, $\bar\alpha_t$, and $\sigma_t$ are predetermined coefficients, and $\bm{\mathrm{z}}\sim\mathcal{N}(\bm{0}, \bm{I})$.

In our method, besides the sampling of $\bm{x}_{t-1}$, we also generate a precise variant $\bm{x}_{t-1}^{(g)}$ by combining 
 the real guide image $\bm{x}^{(g)}_0$ and $\bm{\mathrm{z}}$:
\begin{equation}
    \bm{x}_{t-1}^{(g)} = \frac{1}{\sqrt{\alpha_t}} \bm{x}^{(g)}_0 + \sigma_t \bm{\mathrm{z}}.
\end{equation}
Then similar to classifier-free guidance \cite{ho2021classifier}, the $\bm{x}_{t-1}$ is perturbed by its difference to the guided $\bm{x}_{t-1}^{(g)}$, \textit{i.e.},
\begin{equation} \label{eq:image_guidance}
    \tilde{\bm{x}}_{t-1} = \bm{x}_{t-1} + \gamma_t (\bm{x}_{t-1}^{(g)} - \bm{x}_{t-1}),
\end{equation}
where $\gamma_t$ is the timestep-dependent guidance scale ranging within 0 and 1. $\gamma_t$ can exist in both discrete and continuous form. We utilize a sigmoid function $\gamma_t = 1 - \frac{e^{t - i}}{1 + e^{t - i}}$ where $i$ is the image guidance strength, to enable continuous perturbation of the guidance process. The resulting $\tilde{\bm{x}}_{t-1}$ is used to predict noise and sample $\bm{x}_{t-2}$ in the next timestep.

As visualized in Figure~\ref{fig:vis_cmp}, compared to randomly generating images with a fixed prompt using Stable Diffusion~\cite{Rombach_2022_CVPR}, the proposed image guidance method can obtain images very similar to the guide image. However, we find that under this strict pixel-to-pixel guidance, the synthetic images are difficult to have various background scenes (the same green grass background as the examples in the 2nd row of the figure). Consequently, we propose to guide the foreground object only while keeping the flexibility of background.

\textbf{Selective guidance with attention masks.} To achieve selective guidance, the initial step involves identifying foreground and background pixels. In our context, where text-to-image diffusion models generate text-conditioned images by integrating text embeddings into the cross-attention layers of UNet, the cross attentions between text embeddings and image features inherently reveal the locations of foreground pixels. The use of attentions has been explored in various image editing papers \cite{hertz2022prompt,han2023svdiff,chen2023training,parmar2023zero,liu2023video,chefer2023attend,cao2023masactrl}. In our work, we employ a classical method \cite{chefer2023attend}  to derive attention masks specific to the class.

With the attention mask $\bm{m}_t$ of shape $(1, H, W)$, where $H$ and $W$ denote the height and width of the latent variable $\bm{x}$, respectively, our image guidance generation in Equation (\ref{eq:image_guidance}) is reformulated as
\begin{equation} \label{eq:aig}
    \tilde{\bm{x}}_{t-1} = \bm{x}_{t-1} + \bm{m}_t \odot \gamma_t (\bm{x}_{t-1}^{(g)} - \bm{x}_{t-1}),
\end{equation}
where $\odot$ represents Hadamard (element-wise) product. The procedure of our attentive image guidance is illustrated in Figure~\ref{fig:image_guidance}.

\subsection{Gradient-based Guidance}

The attentive image guidance, as defined in Equation (\ref{eq:aig}), serves as an objective to minimize the disparity between the generated image and the guide image. To enhance the diversity of the generation process, we delve into the untapped potential of synthetic images by introducing a novel guidance mechanism — gradient-based guidance. This mechanism can control more complex and specific generation demands through the design of losses on the image latents.

In text-to-image diffusion models, the text embedding $\bm{c}$ plays a crucial role in image generation. To exert control over the generation process using losses, we propose updating $\bm{c}$ at every timestep through a one-step gradient descent. Before delving into the update mechanism, we first introduce two types of losses, each of which is designed to control the generation with distinct objectives.

\textbf{Contrastive loss.} In Figure~\ref{fig:vis_cmp}, we observe that when using the same class of guide images, the generated images can be highly similar, leading to the redundancy of generated images. Consequently, some generations may not contain sufficient new information than previously generated ones to get further improvements. 
To mitigate the risk of synthetic images being too similar to previously generated ones, we introduce a contrastive loss to encourage the discrepancy between the current generated latent and those generated previously. Inspired by contrastive learning approaches \cite{wu2018unsupervised,he2020momentum}, we incorporate a memory bank to store the latents of all generated images. The contrastive loss measures the distance between the current generation $\bm{x}_t$ and the memory bank corresponding to its class $\bm{B}^{(c)} \in \mathbb{R}^{N\times C\times H \times W}$ (where we sample a maximum of $N=1024$ latents from the bank for efficiency), \textit{i.e.},
\begin{equation}
    \mathcal{L}_{contra} = \frac1N \sum_{i=1}^N max\left(\rho - d(\bm{x}_t, \bm{B}^{(c)}_i), 0\right).
\end{equation}
Here, $d$ represents a distance metric used to measure the distance between two vectors. In this paper, we employ the Euclidean distance as our chosen metric, although another common choice is the KL divergence. The hyper-parameter $\rho=200$ signifies the margin. Specifically, when the distance between the current latent and the previous latent in the memory bank is smaller than $\rho$, it indicates similarity, and a penalty is applied to $\bm{x}_t$ to increase the distance.

As depicted in the bottom row of Figure~\ref{fig:vis_cmp}, our generation approach with the contrastive loss demonstrates a notable increase in the diversity of generated images.

\textbf{Adversarial samples.} To exert further control over the difficulty of generation, we introduce a negative classification loss aimed at increasing the classification difficulty of synthetic images in accordance with the current model. 
Our approach involves denoising $\bm{x}_t$ to cleaned latent $\bm{x}_0$ with the diffusion model, then obtaining the original image $\bm{o}_t$. Subsequently, this image is fed into our current classification model $\Omega$ to obtain predicted logits $\Omega(\bm{o}_t)$, which are then used to compute the negative cross-entropy (CE) loss with the corresponding label $y$, \textit{i.e.},
\begin{equation}
    \mathcal{L}_{adv} = -\mathrm{CE}\left(\Omega(\bm{o}_t), y\right).
\end{equation}

By minimizing the negative cross-entropy loss (\textit{i.e.}, maximizing cross-entropy loss), our method focuses on refining the text embeddings to increase the difficulty of generated images, resulting in observable changes to the image rather than merely adding noise, as shown in Figure \ref{fig:vis_adv}.

\textbf{Gradient update of text embeddings.} With the overall loss $\mathcal{L} = \mathcal{L}_{contra} + \lambda\mathcal{L}_{adv}$, where $\lambda$ is the factor to balance the loss strengths, the text embedding $\bm{c}_{t-1}$ for the next timestep is updated with the normalized gradient:
\begin{equation}
    \bm{c}_{t-1} = \bm{c}_{t} - \nu\frac{\nabla_{\bm{c}_t}\mathcal{L}}{||\nabla_{\bm{c}_t}\mathcal{L}||_2},
\end{equation}
where $\nu$ denotes the learning rate.

\begin{table*}[t]
    \centering
    \small
    \setlength{\tabcolsep}{1.2mm}
    \renewcommand{\arraystretch}{1.1}
    \caption{Top-1 accuracy on ImageNet classfication. SD denotes randomly generating samples with fixed prompt in our method. $*$: Azizi et al. \cite{azizi2023synthetic} uses Imagen \cite{saharia2022photorealistic}, a generative model that is not open-source and is more powerful than the Stable Diffusion used in other methods. We reimplement Azizi et al. \cite{azizi2023synthetic} in supplementary material for fairer comparisons.}
    \label{tab:imgnet}
    \begin{tabular}{l|cc|ccc|cc}
        \toprule
        Method & Model & \thead{Params\\ (M)} & \thead{\#Real\\ (M)} & \thead{\#Gen\\ (M)} & \thead{\underline{\#Gen}\\ \#Real} & ACC & ACC $\Delta$\\
        \midrule
        Real only & \multirow{4}{*}{ResNet-50 \cite{he2016deep}} & \multirow{4}{*}{26} & 1.28 & 0 & 0\% & 76.39 & -\\
        Azizi et al.$^*$ \cite{azizi2023synthetic} & & & 1.28 & 1.2 & 94\% & 78.17 & +1.78\\
        SD random & & & 1.28 & 0.13 & 10\% & 76.64 & +0.25\\
        ActGen (ours) & & & 1.28 & 0.13 & 10\% & \textbf{78.65} & \textbf{+2.26}\\
        \midrule
        Real only & \multirow{4}{*}{ResNet-152 \cite{he2016deep}} & \multirow{4}{*}{64} & 1.28 & 0 & 0\% & 78.59 & -\\
        Azizi et al.$^*$ \cite{azizi2023synthetic} & & & 1.28 & 1.2 & 94\% & 80.15 & +1.56\\
        SD random & & & 1.28 & 0.13 & 10\% & 79.23 & +0.64\\
        ActGen (ours) & & & 1.28 & 0.13 & 10\% & \textbf{80.87} & \textbf{+2.28}\\
        \midrule
        Real only & \multirow{4}{*}{ViT-S/16 \cite{dosovitskiy2020image}} & \multirow{4}{*}{22} & 1.28 & 0 & 0\% & 79.89 & -\\
        Azizi et al.$^*$ \cite{azizi2023synthetic} & & & 1.28 & 1.2 & 94\% & 81.00 & +1.11\\
        SD random & & & 1.28 & 0.08 & 6\% & 80.23 & +0.34 \\
        ActGen (ours) & & & 1.28 & 0.08 & 6\% & \textbf{81.12} & \textbf{+1.23}\\
        \midrule
        Real only & \multirow{4}{*}{DeiT-B/16 \cite{touvron2021training}} & \multirow{4}{*}{87} & 1.28 & 0 & 0\% & 81.79 & -\\
        Azizi et al.$^*$ \cite{azizi2023synthetic} & & & 1.28 & 1.2 & 94\% & 82.84 & +1.04\\
        SD random & & & 1.28 & 0.08 & 6\% & 82.05 & +0.26 \\
        ActGen (ours) & & & 1.28 & 0.08 & 6\% & \textbf{83.29} & \textbf{+1.50}\\
        \bottomrule
    \end{tabular}
\end{table*}

\begin{table*}[t]
    \centering
    \small
    \setlength{\tabcolsep}{0.5mm}
    \renewcommand{\arraystretch}{1.1}
    \caption{Top-1 accuracy on CIFAR-10 and CIFAR-100 datasets. We run the released codes of Da-Fusion \cite{trabucco2023effective} and Real guidance \cite{he2023is} on CIFAR for comparisons. $\dag$: We implement Azizi et al. \cite{azizi2023synthetic} on Stable Diffusion.}
    \label{tab:cifar}
    \begin{tabular}{l|cc|ccc|cc}
        \toprule
        Method & Model & \thead{Params\\ (M)} & \thead{\#Real\\ (K)} & \thead{\#Gen\\ (K)} & \thead{\underline{\#Gen}\\ \#Real} & C10 ACC & C100 ACC\\
        \midrule
        Real only & \multirow{6}{*}{\thead{ResNet-50\\ \cite{he2016deep}}} & \multirow{6}{*}{24} & 50 & 0 & 0\% & 95.02$\pm$0.17 & 77.06$\pm$0.28\\
        Azizi et al.$^\dag$ \cite{azizi2023synthetic} & & & 50 & 9.6 & 19\% & 95.18$\pm$0.26 & 77.07$\pm$0.45\\
        SD random & & & 50 & 9.6 & 19\% & 95.26$\pm$0.22 & 77.17$\pm$0.24\\
        Da-Fusion \cite{trabucco2023effective} & & & 50 & 9.6 & 19\% & 95.14$\pm$0.36 & 76.26$\pm$0.42\\
        Real guidance \cite{he2023is} & & & 50 & 9.6 & 19\% & 95.22$\pm$0.19 & 76.83$\pm$0.36\\
        ActGen (ours) & & & 50 & 9.6 & 19\% & \textbf{95.53}$\pm$0.37 & \textbf{77.33}$\pm$0.34\\
        \midrule
        Real only & \multirow{6}{*}{\thead{VGG-16\\ \cite{simonyan2014very}}} & \multirow{6}{*}{15} & 50 & 0 & 0\% & 93.73$\pm$0.21 & 73.96$\pm$0.31\\
        Azizi et al.$^\dag$ \cite{azizi2023synthetic} & & & 50 & 9.6 & 19\% & 94.35$\pm$0.39 & 73.84$\pm$0.28\\
        SD random & & & 50 & 9.6 & 19\% & 94.01$\pm$0.37 & 74.13$\pm$0.43\\
        Da-Fusion \cite{trabucco2023effective} & & & 50 & 9.6 & 19\% & 93.97$\pm$0.51 & 73.68$\pm$0.39\\
        Real guidance \cite{he2023is} & & & 50 & 9.6 & 19\% & 94.22$\pm$0.31 & 74.16$\pm$0.37\\
        ActGen (ours) & & & 50 & 9.6 & 19\% & \textbf{94.62}$\pm$0.34 & \textbf{74.47}$\pm$0.35\\
        \bottomrule
    \end{tabular}
\end{table*}
\section{Experiments}

To validate the efficacy of our method sufficiently, we conduct experiments to compare with existing methods on two types of image classification tasks: supervised image classification and few-shot image classification.

\subsection{Implementation Details}

\textbf{Diffusion models.} On supervised image classification, we use Stable Diffusion V2.1 base\footnote{\href{https://huggingface.co/stabilityai/stable-diffusion-2-1-base}{https://huggingface.co/stabilityai/stable-diffusion-2-1-base}} as the model for text-to-image generation. We use a DDPM sampler with $40$ diffusion steps to generate $512\times512$ images. The guidance scale $s$ of classifier-free guidance is set to 15 and the image guidance scale $i$ is set to 12.5. On few-shot learning, GLIDE\footnote{\href{https://github.com/openai/glide-text2im}{https://github.com/openai/glide-text2im}} is used with classifier guidance to align with experiments conducted by \cite{he2023is}. A DDPM sampler with $40$ diffusion steps is used to generate $256\times256$ images with guidance scale $s$ set to 3.

\subsection{Results on Image Classification}
\textbf{Settings.} We conduct experiments on ImageNet and CIFAR datasets to validate our superiority on traditional image classification task. On both ImageNet and CIFAR datasets, we partition a validation set with $10$K images from the train set, and generate 64 images per GPU after each epoch. We conduct generation only on the first half epochs, since the learning rate is small in the subsequent training period and the newly generated images would have small effects on the performance. On ImageNet, we follow the same training strategies in \cite{azizi2023synthetic}; while on CIFAR, a 300-epoch training strategy is adopted. Detailed settings are summarized in Supplementary Material. 
For CIFAR datasets, we train the model 5 trials independently and report their mean and standard deviation on accuracy.

\textbf{Results on ImageNet.} Following \cite{azizi2023synthetic}, we adopt our ActGen on ResNet-50~\cite{he2016deep} and ViT-S/16~\cite{dosovitskiy2020image}. For comparison with generation without guidance, we also implement a \textit{SD random} baseline, which uses the same active generation strategy as our ActGen but only generates images with a fixed prompt on each class. As the results shown in Table~\ref{tab:imgnet}, compared to the conventional training with real images only, our method enjoys significant accuracy improvements, while only has 10\% additional images. Compared to the pioneering method~\cite{azizi2023synthetic}, our method obtains better accuracies while only has less than 10\% images generated. For example, we obtain 78.65\% accuracy on ResNet-50, which outperforms the \textit{real only} baseline by a large margin of 2.26\%, and we also achieve 0.48\% accuracy increment over \cite{azizi2023synthetic} while saving $\sim$1M synthetic images. In contrast, the \textit{SD random} only yields marginal improvements, showing that a more effective and deterministic generation method is important to obtain larger improvements with limited samples. These demonstrate our efficacy on generating valuable images for classification.

\textbf{Results on CIFAR.} To validate our efficacy on smaller and simpler datasets, we also implement our method on CIFAR-10 and CIFAR-100 datasets. As shown in Table~\ref{tab:cifar}, though the models are easily to converge and overfit on the training set, our method still gains obvious improvements by introducing more hard samples. For example, on VGG-16, ActGen achieves significant 0.89 and 0.51 improvements compared to the baseline, with only 9.6K images generated.

\subsection{Results on Few-Shot Image Classification}

\textbf{Settings.} In order to showcase our proficiency in the few-shot learning problem, we conduct experiments on the EuroSAT dataset \cite{helber2019eurosat}. Real images are used just for the purpose of validation, in which the softmax confidence of the validation samples is transformed into image guidance scale. As confidence levels increase, the corresponding guiding scale decreases, so facilitating the introduction of greater diversity. Our generation strategy exclusively uses image guidance (without masking, gradient-base guidance). The dataset is partitioned according to code base\footnote{\href{https://github.com/saic-fi/Bayesian-Prompt-Learning/tree/main}{https://github.com/saic-fi/Bayesian-Prompt-Learning/tree/main}} provided by \cite{derakhshani2023variational}. Detailed training settings are in Supplementary Material.

\textbf{Results on EuroSAT.} We replicate the experiments conducted by \cite{he2023is} on EuroSAT few-shot image classification, employing the best strategy as our baseline with 2,000 samples for matching with our generated amount.  From Table~\ref{tab:few_shot_eurosat}, with only 25\% of sample quantity of baseline, we observe a significant improvement in performance across all shots when comparing to the 2K baseline. Moreover, even when comparing our 2K generation with 8K generation of \cite{he2023is}, ActGen can also achieve performance gains of 0.25\%, 0.02\%, 1.6\%, and 0.85\% for 16, 4, 2, and 1 shot(s), respectively.

\begin{table}[t]
    \centering
    \small
    \setlength{\tabcolsep}{1.1mm}
    \renewcommand{\arraystretch}{1.1}
    \caption{Accuracy of few-shot classification on EuroSAT.}
    \label{tab:few_shot_eurosat}
    \begin{tabular}{l|c|ccccc}
        \toprule
        Method & \#Gen & 1-shot & 2-shot & 4-shot & 8-shot & 16-shot\\
        \midrule
        Real guidance \cite{he2023is} & 2K & 66.00 & 75.72 & 81.35 & 81.51 & 86.27\\
        Real filtering \cite{he2023is} & 2K & \textbf{70.32} & 76.63 & 80.63 & 81.43 & 85.23\\
        Da-Fusion \cite{trabucco2023effective} & 2K & 53.44 & 64.64 & 66.90 & 75.21 & 80.96\\
        ActGen (ours) & 2K & 66.78 & \textbf{77.81} & \textbf{81.77} & \textbf{81.51} & \textbf{87.25}\\
        \bottomrule
    \end{tabular}
\end{table}

\subsection{Ablation Study}

\textbf{Effect of different guidance mechanisms.} In Figure~\ref{fig:vis_cmp} and Figure~\ref{fig:vis_adv}, we compare the differences of the generated images with different guidance mechanisms in our method. Now we conduct experiments to make the numerical comparisons on the final accuracy of them. As shown in Table~\ref{tab:ab_guidance}, all of our guidance mechanisms contribute to performance increment due to the better diversity and generation quality.

\textbf{Ablation study on numbers of validation images.} The validation set play a crucial role in identifying hard samples for generation. Here we conduct experiments to show the influence of its size to the performance. As shown in Figure~\ref{fig:num_val}, with only 0.1K and 1K images, the validation set is not sufficient to cover all the classes in ImageNet, thus leading to a relative poor performance. When the size becomes larger than 5K, its performance tends to be stable.

\begin{table}[t]
    \centering
    \begin{minipage}[c]{0.48\textwidth}
        \small
        \setlength{\tabcolsep}{0.8mm}
        \renewcommand{\arraystretch}{1.1}
        \caption{Ablation study on guidance mechanisms.}
        \label{tab:ab_guidance}
        \begin{tabular}{c|cccc|c}
            \toprule
            Random & IG & AIG & $\mathcal{L}_{contra}$ & $\mathcal{L}_{adv}$ & ACC\\
            \midrule
            &&&&& 76.39\\
            \checkmark &&&&& 76.64\\
            & \checkmark &  &  &  & 77.93\\
            & \checkmark & \checkmark &  &  & 78.15\\
            & \checkmark & \checkmark & \checkmark &  &  78.36\\
            & \checkmark & \checkmark & \checkmark & \checkmark & \textbf{78.65}\\
            \bottomrule
        \end{tabular}
    \end{minipage}\hfill
    \begin{minipage}[c]{0.48\textwidth}
        {\vspace{4mm}\includegraphics[width=1.0\linewidth]{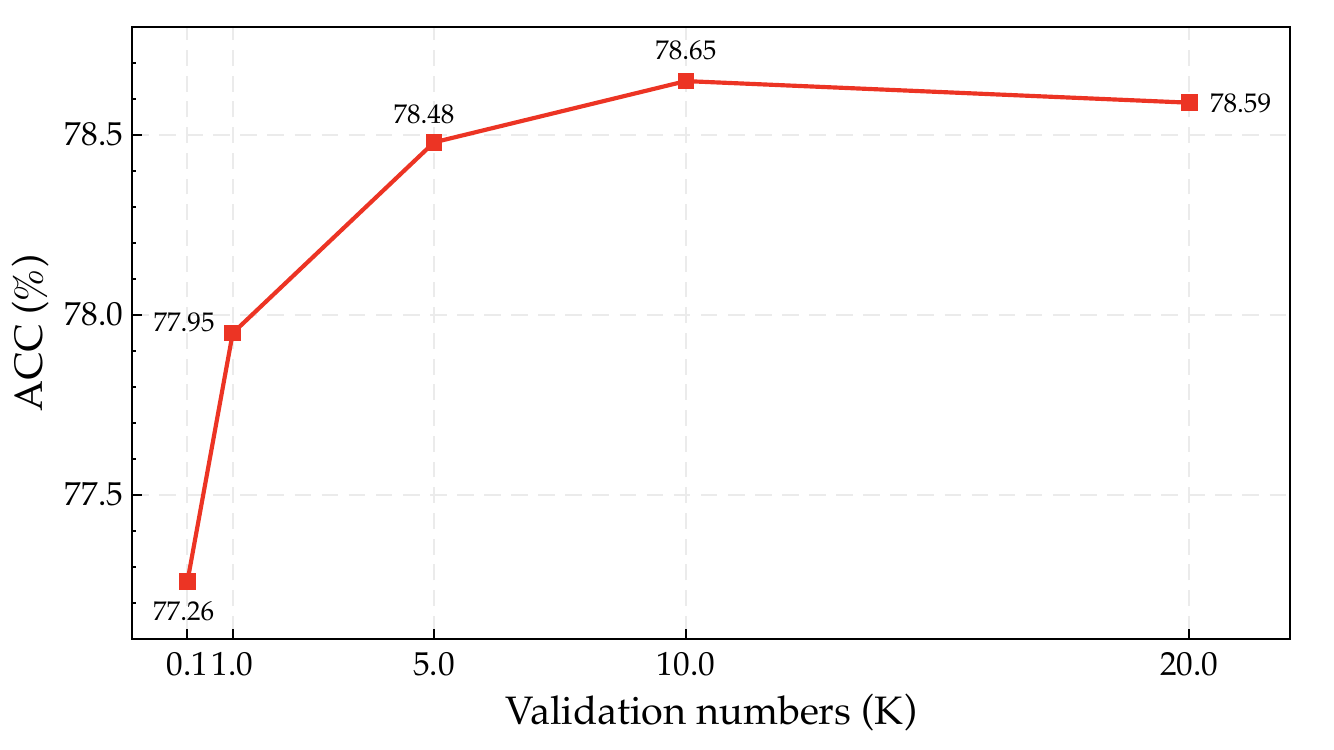}}
	    \makeatletter\def\@captype{figure}\makeatother
        \caption{Accuracy of ResNet-50 on ImageNet with different numbers of validation images.}
        \label{fig:num_val}
    \end{minipage}
\end{table}

\textbf{Training cost analysis.} In ActGen, the generation of hard images and training of them introduce additional computational cost. Taking the training of ResNet-50 on ImageNet as an example. the traditional training with real images takes 9.6 GPU days on 32 NVIDIA V100 GPUs. While for ActGen, it takes 4.5 GPU days for generating images, resulting in 15.2 GPU days of total training time. However, compared to generating $10\times$ of samples in previous method, which would cost $\sim$40 GPU days for generation and $2\times$ time for training, our additional training cost is acceptable. Meanwhile, comparing with Stable Diffusion v1.5, ActGen increases its VRAM of from 8.4G to 8.7G. Additionally, the generation times are comparable, with SD at 2.8 s/img and ActGen at 3.0 s/img.

More experiments are in Supplementary Material.

\section{Conclusion}

In this study, we address the efficiency challenges associated with image generation in the context of enhancing image classification accuracy using deep generative models. ActGen adopts a training-aware approach inspired by active learning principles, focusing on generating images that mimic challenging or misclassified samples encountered by the model. By incorporating these images into the training set, ActGen significantly improves model performance. Key innovations include an attentive image guidance technique within the denoising process, preserving similar foreground objects while diversifying the background. Additionally, a gradient-based generation guidance method generates more challenging samples, avoiding excessive similarity to previously generated images. Experimental results on CIFAR and ImageNet demonstrate ActGen's competitive performance with a notably reduced number of generated images. This work represents a promising step toward more efficient and practical deep generative models for image classification.

\section*{Acknowledgements}
This work was supported in part by the Australian Research Council under Projects DP210101859 and FT230100549.

%
%
\bibliographystyle{splncs04}
\bibliography{main}

\begin{thebibliography}{10}
\providecommand{\url}[1]{\texttt{#1}}
\providecommand{\urlprefix}{URL }
\providecommand{\doi}[1]{https://doi.org/#1}

\bibitem{azizi2023synthetic}
Azizi, S., Kornblith, S., Saharia, C., Norouzi, M., Fleet, D.J.: Synthetic data from diffusion models improves imagenet classification. arXiv preprint arXiv:2304.08466  (2023)

\bibitem{bengio2009curriculum}
Bengio, Y., Louradour, J., Collobert, R., Weston, J.: Curriculum learning. In: Proceedings of the 26th annual international conference on machine learning. pp. 41--48 (2009)

\bibitem{besnier2020dataset}
Besnier, V., Jain, H., Bursuc, A., Cord, M., P{\'e}rez, P.: This dataset does not exist: training models from generated images. In: ICASSP 2020-2020 IEEE International Conference on Acoustics, Speech and Signal Processing (ICASSP). pp.~1--5. IEEE (2020)

\bibitem{brock2018large}
Brock, A., Donahue, J., Simonyan, K.: Large scale gan training for high fidelity natural image synthesis. arXiv preprint arXiv:1809.11096  (2018)

\bibitem{cao2023masactrl}
Cao, M., Wang, X., Qi, Z., Shan, Y., Qie, X., Zheng, Y.: Masactrl: Tuning-free mutual self-attention control for consistent image synthesis and editing. arXiv preprint arXiv:2304.08465  (2023)

\bibitem{carion2020end}
Carion, N., Massa, F., Synnaeve, G., Usunier, N., Kirillov, A., Zagoruyko, S.: End-to-end object detection with transformers. In: European conference on computer vision. pp. 213--229. Springer (2020)

\bibitem{chefer2023attend}
Chefer, H., Alaluf, Y., Vinker, Y., Wolf, L., Cohen-Or, D.: Attend-and-excite: Attention-based semantic guidance for text-to-image diffusion models. ACM Transactions on Graphics (TOG)  \textbf{42}(4),  1--10 (2023)

\bibitem{chen2018encoder}
Chen, L.C., Zhu, Y., Papandreou, G., Schroff, F., Adam, H.: Encoder-decoder with atrous separable convolution for semantic image segmentation. In: Proceedings of the European conference on computer vision (ECCV). pp. 801--818 (2018)

\bibitem{chen2023training}
Chen, M., Laina, I., Vedaldi, A.: Training-free layout control with cross-attention guidance. arXiv preprint arXiv:2304.03373  (2023)

\bibitem{derakhshani2023variational}
Derakhshani, M.M., Sanchez, E., Bulat, A., da~Costa, V.G.T., Snoek, C.G., Tzimiropoulos, G., Martinez, B.: Bayesian prompt learning for image-language model generalization. ICCV  (2023)

\bibitem{dhariwal2021diffusion}
Dhariwal, P., Nichol, A.: Diffusion models beat gans on image synthesis. Advances in neural information processing systems  \textbf{34},  8780--8794 (2021)

\bibitem{dosovitskiy2020image}
Dosovitskiy, A., Beyer, L., Kolesnikov, A., Weissenborn, D., Zhai, X., Unterthiner, T., Dehghani, M., Minderer, M., Heigold, G., Gelly, S., et~al.: An image is worth 16x16 words: Transformers for image recognition at scale. In: International Conference on Learning Representations (2020)

\bibitem{goodfellow2014generative}
Goodfellow, I., Pouget-Abadie, J., Mirza, M., Xu, B., Warde-Farley, D., Ozair, S., Courville, A., Bengio, Y.: Generative adversarial nets. Advances in neural information processing systems  \textbf{27} (2014)

\bibitem{guo2020positive}
Guo, T., Xu, C., Huang, J., Wang, Y., Shi, B., Xu, C., Tao, D.: On positive-unlabeled classification in gan. In: Proceedings of the ieee/cvf conference on computer vision and pattern recognition. pp. 8385--8393 (2020)

\bibitem{han2023svdiff}
Han, L., Li, Y., Zhang, H., Milanfar, P., Metaxas, D., Yang, F.: Svdiff: Compact parameter space for diffusion fine-tuning. arXiv preprint arXiv:2303.11305  (2023)

\bibitem{he2020momentum}
He, K., Fan, H., Wu, Y., Xie, S., Girshick, R.: Momentum contrast for unsupervised visual representation learning. In: Proceedings of the IEEE/CVF conference on computer vision and pattern recognition. pp. 9729--9738 (2020)

\bibitem{he2016deep}
He, K., Zhang, X., Ren, S., Sun, J.: Deep residual learning for image recognition. In: Proceedings of the IEEE conference on computer vision and pattern recognition. pp. 770--778 (2016)

\bibitem{he2023is}
He, R., Sun, S., Yu, X., Xue, C., Zhang, W., Torr, P., Bai, S., Qi, X.: Is synthetic data from generative models ready for image recognition? In: International Conference on Learning Representations (2023), \url{https://openreview.net/forum?id=nUmCcZ5RKF}

\bibitem{helber2019eurosat}
Helber, P., Bischke, B., Dengel, A., Borth, D.: Eurosat: A novel dataset and deep learning benchmark for land use and land cover classification. IEEE Journal of Selected Topics in Applied Earth Observations and Remote Sensing  \textbf{12}(7),  2217--2226 (2019)

\bibitem{hertz2022prompt}
Hertz, A., Mokady, R., Tenenbaum, J., Aberman, K., Pritch, Y., Cohen-Or, D.: Prompt-to-prompt image editing with cross attention control. arXiv preprint arXiv:2208.01626  (2022)

\bibitem{ho2020denoising}
Ho, J., Jain, A., Abbeel, P.: Denoising diffusion probabilistic models. Advances in Neural Information Processing Systems  \textbf{33},  6840--6851 (2020)

\bibitem{ho2021classifier}
Ho, J., Salimans, T.: Classifier-free diffusion guidance. In: NeurIPS 2021 Workshop on Deep Generative Models and Downstream Applications (2021)

\bibitem{karras2019style}
Karras, T., Laine, S., Aila, T.: A style-based generator architecture for generative adversarial networks. In: Proceedings of the IEEE/CVF conference on computer vision and pattern recognition. pp. 4401--4410 (2019)

\bibitem{lin2017focal}
Lin, T.Y., Goyal, P., Girshick, R., He, K., Doll{\'a}r, P.: Focal loss for dense object detection. In: Proceedings of the IEEE international conference on computer vision. pp. 2980--2988 (2017)

\bibitem{liu2023video}
Liu, S., Zhang, Y., Li, W., Lin, Z., Jia, J.: Video-p2p: Video editing with cross-attention control. arXiv preprint arXiv:2303.04761  (2023)

\bibitem{long2015fully}
Long, J., Shelhamer, E., Darrell, T.: Fully convolutional networks for semantic segmentation. In: Proceedings of the IEEE conference on computer vision and pattern recognition. pp. 3431--3440 (2015)

\bibitem{loshchilov2015online}
Loshchilov, I., Hutter, F.: Online batch selection for faster training of neural networks. arXiv preprint arXiv:1511.06343  (2015)

\bibitem{nichol2021glide}
Nichol, A., Dhariwal, P., Ramesh, A., Shyam, P., Mishkin, P., McGrew, B., Sutskever, I., Chen, M.: Glide: Towards photorealistic image generation and editing with text-guided diffusion models. arXiv preprint arXiv:2112.10741  (2021)

\bibitem{nichol2021improved}
Nichol, A.Q., Dhariwal, P.: Improved denoising diffusion probabilistic models. In: International Conference on Machine Learning. pp. 8162--8171. PMLR (2021)

\bibitem{parkhi2012cats}
Parkhi, O.M., Vedaldi, A., Zisserman, A., Jawahar, C.: Cats and dogs. In: 2012 IEEE conference on computer vision and pattern recognition. pp. 3498--3505. IEEE (2012)

\bibitem{parmar2023zero}
Parmar, G., Kumar~Singh, K., Zhang, R., Li, Y., Lu, J., Zhu, J.Y.: Zero-shot image-to-image translation. In: ACM SIGGRAPH 2023 Conference Proceedings. pp. 1--11 (2023)

\bibitem{ravuri2019seeing}
Ravuri, S., Vinyals, O.: Seeing is not necessarily believing: Limitations of biggans for data augmentation  (2019)

\bibitem{ren2021survey}
Ren, P., Xiao, Y., Chang, X., Huang, P.Y., Li, Z., Gupta, B.B., Chen, X., Wang, X.: A survey of deep active learning. ACM computing surveys (CSUR)  \textbf{54}(9),  1--40 (2021)

\bibitem{ren2015faster}
Ren, S., He, K., Girshick, R., Sun, J.: Faster r-cnn: Towards real-time object detection with region proposal networks. Advances in neural information processing systems  \textbf{28} (2015)

\bibitem{Rombach_2022_CVPR}
Rombach, R., Blattmann, A., Lorenz, D., Esser, P., Ommer, B.: High-resolution image synthesis with latent diffusion models. In: Proceedings of the IEEE/CVF Conference on Computer Vision and Pattern Recognition (CVPR). pp. 10684--10695 (June 2022)

\bibitem{saharia2022photorealistic}
Saharia, C., Chan, W., Saxena, S., Li, L., Whang, J., Denton, E.L., Ghasemipour, K., Gontijo~Lopes, R., Karagol~Ayan, B., Salimans, T., et~al.: Photorealistic text-to-image diffusion models with deep language understanding. Advances in Neural Information Processing Systems  \textbf{35},  36479--36494 (2022)

\bibitem{schaul2015prioritized}
Schaul, T., Quan, J., Antonoglou, I., Silver, D.: Prioritized experience replay. arXiv preprint arXiv:1511.05952  (2015)

\bibitem{shi2020improving}
Shi, Z., Zhou, X., Qiu, X., Zhu, X.: Improving image captioning with better use of captions. arXiv preprint arXiv:2006.11807  (2020)

\bibitem{shipard2023diversity}
Shipard, J., Wiliem, A., Thanh, K.N., Xiang, W., Fookes, C.: Diversity is definitely needed: Improving model-agnostic zero-shot classification via stable diffusion. In: Proceedings of the IEEE/CVF Conference on Computer Vision and Pattern Recognition. pp. 769--778 (2023)

\bibitem{simonyan2014very}
Simonyan, K., Zisserman, A.: Very deep convolutional networks for large-scale image recognition. arXiv preprint arXiv:1409.1556  (2014)

\bibitem{sohl2015deep}
Sohl-Dickstein, J., Weiss, E., Maheswaranathan, N., Ganguli, S.: Deep unsupervised learning using nonequilibrium thermodynamics. In: International conference on machine learning. pp. 2256--2265. PMLR (2015)

\bibitem{song2020carpe}
Song, H., Kim, M., Kim, S., Lee, J.G.: Carpe diem, seize the samples uncertain" at the moment" for adaptive batch selection. In: Proceedings of the 29th ACM International Conference on Information \& Knowledge Management. pp. 1385--1394 (2020)

\bibitem{touvron2021training}
Touvron, H., Cord, M., Douze, M., Massa, F., Sablayrolles, A., J{\'e}gou, H.: Training data-efficient image transformers \& distillation through attention. In: International conference on machine learning. pp. 10347--10357. PMLR (2021)

\bibitem{trabucco2023effective}
Trabucco, B., Doherty, K., Gurinas, M., Salakhutdinov, R.: Effective data augmentation with diffusion models. arXiv preprint arXiv:2302.07944  (2023)

\bibitem{wu2018unsupervised}
Wu, Z., Xiong, Y., Yu, S.X., Lin, D.: Unsupervised feature learning via non-parametric instance discrimination. In: Proceedings of the IEEE conference on computer vision and pattern recognition. pp. 3733--3742 (2018)

\bibitem{zhang2021datasetgan}
Zhang, Y., Ling, H., Gao, J., Yin, K., Lafleche, J.F., Barriuso, A., Torralba, A., Fidler, S.: Datasetgan: Efficient labeled data factory with minimal human effort. In: Proceedings of the IEEE/CVF Conference on Computer Vision and Pattern Recognition. pp. 10145--10155 (2021)

\bibitem{zhao2017pyramid}
Zhao, H., Shi, J., Qi, X., Wang, X., Jia, J.: Pyramid scene parsing network. In: Proceedings of the IEEE conference on computer vision and pattern recognition. pp. 2881--2890 (2017)

\bibitem{zhou2023training}
Zhou, Y., Sahak, H., Ba, J.: Training on thin air: Improve image classification with generated data. arXiv preprint arXiv:2305.15316  (2023)

\end{thebibliography}

\newpage
\appendix

\section{Implementations}

\subsection{Training Strategies on ImageNet}

For fair comparison, we follow the training strategies of ResNet-50 and ViT-S/16 in \cite{azizi2023synthetic}. Detailed strategies are summarized in Table~\ref{tab:strategy_imgnet}.

\begin{table}[h]
    \centering
    \small
    \setlength{\tabcolsep}{1.5mm}
    \renewcommand{\arraystretch}{1.1}
    \caption{Training strategies on ImageNet.}
    \label{tab:strategy_imgnet}
    \begin{tabular}{c|cc}
        \toprule
        Model & ResNet-50 & ViT-S/16\\
        \midrule
        Batch size & 4096 & 1024\\
        Optimizer & Momentum SGD & AdamW\\
        Learning rate & 1.6 & 0.001\\
        Decay method & Cosine & Cosine\\
        Weight decay & 1e-4 & 1e-4\\
        Warmup epochs & 5 & 10\\
        Label smoothing & 0.1 & 0.1\\
        Dropout rate & 0.25 & -\\
        Rand Augment & 10 & 10\\
        Mixup prob. & - & 0.2\\
        Cutmix prob. & - & 1.0\\
        \bottomrule
    \end{tabular}
\end{table}

\subsection{Training Strategy on CIFAR}

On CIFAR-10 and CIFAR-100 datasets, we use a baseline strategy for all the models, as shown in Table~\ref{tab:strategy_cifar}. The data augmentations utilized in training are random cropping and random horizontal flip.

\begin{table}[h]
    \centering
    \small
    \setlength{\tabcolsep}{1.5mm}
    \renewcommand{\arraystretch}{1.1}
    \caption{Training strategy on CIFAR.}
    \label{tab:strategy_cifar}
    \begin{tabular}{c|cccccc}
        \toprule
        Model & ResNet-50\\
        \midrule
        Batch size & 128\\
        Optimizer & Momentum SGD\\
        Learning rate & 0.1\\
        Decay method & Cosine\\
        Weight decay & 1e-4\\
        \bottomrule
    \end{tabular}
\end{table}

\subsection{Training strategies on few-shot classification}

We follow the code\footnote{\href{https://github.com/saic-fi/Bayesian-Prompt-Learning/tree/main}{https://github.com/saic-fi/Bayesian-Prompt-Learning/tree/main}} provided by \cite{derakhshani2023variational} to partition EuroSAT dataset. The few-shot real images are randomly sampled from training set and used for validation only during generation phase. For N-shot-M-way problem, we generate N (In this experiment, N is number of shots * batch size, batch size is 1 for EuroSAT and 2 for Pets) images each generation epoch with a total of $\frac{2,000}{N}$ epochs. During the transition between each generation epoch, the model undergoes a process of fine-tuning for a duration of 10 epochs, with a learning rate of 0.001. The generation employs a confidence-to-guidance conversion function $\eta_f = \frac{L}{1 + e^{k^{(f - u)}}} + p$, where $L = 30, k = 10, p = 5, u = [0.1, 0.15, 0.5, 0.9, 1.1]$ respectively for [1, 2, 4, 8, 16] shot(s). The zero-shot CLIP-ResNet50 model is fine-tuned using a mix-training technique \cite{he2023is} for a total of 100 epochs with learning rate of 0.01 after the generating process.

\subsection{Generation Details}
We use the pretrained Stable Diffusion V2.1 base model for text-to-image generation on ImageNet and CIFAR datasets. We use the method in \cite{chefer2023attend} to get the attention mask in our image guidance. We use a probability of $\frac{0.5\times\mathrm{epoch}}{\mathrm{total\_epoch}}$ to generate adversarial samples or non-adversarial samples otherwise, where this increasing probability is inspired by the curriculum learning~\cite{bengio2009curriculum}, which states that the optimization should gradually increase its learning difficulty for the model. The gradient-based guidance is utilized in the first 10 iterations out of the total 40 iterations in DDPM scheduler. This is to reduce the computation and memory cost and ensure high-fidelity generations.

\subsection{Training Procedure of ActGen}

In Algorithm~\ref{alg:actgen}, we summarize our algorithm and show our active generation process during training.

\begin{algorithm}
\caption{Active training procedure of ActGen}
\label{alg:actgen}
\begin{algorithmic}[1]
    \REQUIRE Generation model $G$, classification model $\Omega$, training dataset $\mathcal{D}_{tr}$.
    \STATE $\mathcal{D}_{tr}, \mathcal{D}_{val} \leftarrow \mathrm{Partition}(\mathcal{D}_{tr}$); \hfill \graytext{\textit{\# partition train set to train and val sets}}
    \FOR{epoch in total\_epoch}
        \STATE $\mathrm{Train}(\mathcal{D}_{tr}, \Omega)$; \hfill\graytext{\textit{\# train $\Omega$ for one epoch}}
        \STATE $X_{hard} \leftarrow \mathrm{Val}(\mathcal{D}_{val}, \Omega)$; \hfill \graytext{\textit{\# identify hard samples from $\mathcal{D}_{val}$}}
        \STATE $X_{gen} \leftarrow \mathrm{Gen}(X_{hard}, G, \Omega)$; \hfill \graytext{\textit{\# generate images}}
        \STATE $\mathcal{D}_{tr} \leftarrow \mathcal{D}_{tr} \cup X_{gen}$; \hfill\graytext{\textit{\# extend $X_{gen}$ to $\mathcal{D}_{tr}$}}
    \ENDFOR
    \ENSURE Trained model $\Omega$.
\end{algorithmic} 
\end{algorithm}

\section{More Experiments}

\subsection{Few-shot Classification on Pets Dataset}

Our few-shot generation strategies have shown efficiency and effectiveness on the EuroSAT dataset which has low zero-shot classification accuracy of 38.31\%. To further validate the viability of our method, we conducted experiments on the Pets dataset \cite{parkhi2012cats}, which has high zero-shot accuracy of 85.72\%. This finding illustrates that our methodology continues to be efficacious even in situations when the incorporation of synthetic data offers limited potential for enhancing performance.

The adverse impact on classification performance is evident as the quantity of synthetic images generated using the approach proposed by \cite{he2023is} increases on Pets, as demonstrated in Table~\ref{tab:num_samples}. With around 1,850 samples, it is probable that the model has attained its optimal performance. Therefore, we have chosen the generation number of 1,850 for our strategy.

\begin{table}[h]
    \centering
    \small
    \setlength{\tabcolsep}{2mm}
    \renewcommand{\arraystretch}{1.1}
    \caption{Few-shot classification accuracy with respect to the number of synthetic samples on Pets dataset.}
    \vspace{-2mm}
    \label{tab:num_samples}
\begin{tabular}{c|ccccc}
\toprule
\#Gen & \multicolumn{1}{c}{16 shots}  & \multicolumn{1}{c}{8 shots}            & \multicolumn{1}{c}{4 shots}   & \multicolumn{1}{c}{2 shots}            & \multicolumn{1}{c}{1 shot}             \\ \midrule
29,600     &  89.97 & 88.33                                  & 87.79          & 87.54                                  & 87.41                                  \\
22,200     & 89.94 & 88.42                                  & 87.98          & 87.54                                  & 87.57                                  \\
11,100     & 89.94 & 88.42                                  & 87.78 & 87.52                                  & 87.57                                  \\
7,400      &        \textbf{90.00}                       & 88.36                                  &                    87.82           & 87.54                                  & 87.63                                  \\
3,700      &             89.88                  & 88.42                                  &                       87.92        & 87.63                                  & 87.63                                  \\
1,850      &             89.72                 & \cellcolor[HTML]{FFFFFF}\textbf{88.52} &                     \textbf{87.98}          & \cellcolor[HTML]{FFFFFF}\textbf{87.73} & \cellcolor[HTML]{FFFFFF}\textbf{87.65} \\
1,480      &               89.62                & 88.47                                  &                     87.93          & 87.72                                  & 87.63                                  \\
1,110      &              89.15                 & 88.28                                  &                   87.93            & 87.63                                  & 87.65                                  \\
740        &              89.04                 & 88.24                                  &                      87.89         & 87.59                                  & 87.33                                  \\
370        &               88.55                & 88.14                                  &                     87.74          & 87.41                                  & 87.27                                  \\ \bottomrule
\end{tabular}
\end{table}

As seen in Table~\ref{tab:pets}, our few-shot approach on the Pets dataset continues to exhibit both efficiency and effectiveness, achieving optimal performance by utilizing just 40\% of the synthetic data quantity compared to the method proposed by \cite{he2023is}. Our approach demonstrates comparable performance compared to the previous method with slight increments of 0.05\%, 0.06\%, 0.03\%, 0.17\% and 0.22\% on 16, 8, 4, 2, 1 shot(s).

\begin{table}[t]
\centering
    \small
    \setlength{\tabcolsep}{1.2mm}
    \renewcommand{\arraystretch}{1.1}
    \caption{Accuracy of few-shot classification on Pets. The best accuracy is marked as \textbf{bold}. The second best accuracy which is better than baseline is marked as \underline{underline}, respectively.}
    \label{tab:pets}
\begin{tabular}{
>{\columncolor[HTML]{FFFFFF}}c |
>{\columncolor[HTML]{FFFFFF}}l |
>{\columncolor[HTML]{FFFFFF}}c 
>{\columncolor[HTML]{FFFFFF}}c 
>{\columncolor[HTML]{FFFFFF}}c 
>{\columncolor[HTML]{FFFFFF}}c 
>{\columncolor[HTML]{FFFFFF}}c }
\toprule
Method                                                  & \#Gen & 16-shot        & 8-shot         & 4-shot         & 2-shot         & 1-shot         \\ \hline
\cellcolor[HTML]{FFFFFF}                                & 29.6K &         89.97       & 88.33          &         87.79       & 87.54          & 87.41          \\
\multirow{-2}{*}{\cellcolor[HTML]{FFFFFF}\cite{he2023is}}            & 1.85K &         89.72       & 88.52          &         87.98       & 87.73          & 87.65          \\ 
~ & 0.74K & 89.04 & 88.24 & 87.59 & 87.59 & 87.33\\
\midrule
\cellcolor[HTML]{FFFFFF}                                & 1.85K & \textbf{90.05} & \textbf{88.58} & \textbf{88.01} & \textbf{87.90} & \textbf{87.87} \\
\cellcolor[HTML]{FFFFFF}                                & 1.48K & 89.62          & \underline{88.53}          & 87.93          & \underline{87.76}          & \underline{87.82}          \\
\cellcolor[HTML]{FFFFFF}                                & 1.11K & 89.32          & \underline{88.53}          & 87.82          & \underline{87.76}          & \underline{87.79}          \\
\multirow{-4}{*}{\cellcolor[HTML]{FFFFFF}ActGen (ours)} & 0.74K & 89.04          & 88.25          & 87.79          & \underline{87.76}          & \underline{87.68}          \\ \bottomrule
\end{tabular}
\end{table}

\subsection{Compare with Azizi et al. \cite{azizi2023synthetic} on Stable Diffusion}

One of our prior work, Azizi et al. \cite{azizi2023synthetic}, which generates images on ImageNet classes to improve the classification performance, leveraged the powerful Imagen \cite{saharia2022photorealistic} generative model. However, the model is not publicly available. For a fairer comparison, we implement Azizi et al. \cite{azizi2023synthetic} on ImageNet with Stable Diffusion and generate the same number of images as our ActGen. As the results reported in Table \ref{tab:ab_cmp_azizi}, \cite{azizi2023synthetic} on Stable Diffusion performs worse than the origin on Imagen, while our ActGen outperforms all of them.

\begin{table}[h]
    \centering
    \small
    \setlength{\tabcolsep}{2mm}
    \renewcommand{\arraystretch}{1.2}
    \caption{Accuracy of ImageNet classification. SD: we implement Azizi et al. \cite{azizi2023synthetic} on Stable Diffusion and generate the same number of images as our ActGen.}
    \label{tab:ab_cmp_azizi}
    \begin{tabular}{l|cccc}
        \toprule
        Model & Real only & Azizi et al. \cite{azizi2023synthetic} & Azizi et al. \cite{azizi2023synthetic} (SD) & ActGen (ours)\\
        \midrule
        ResNet-50 & 76.39 & 78.17 & 76.83 & \textbf{78.65}\\
        ViT-S/16 & 79.89 & 81.00 & 80.41 & \textbf{81.12}\\
        \bottomrule
    \end{tabular}
\end{table}

We also compared our method with traditional training, original SD, and Azizi et al. \cite{azizi2023synthetic} (excluding its cost of training generative model). As shown in Table~\ref{tab:ab_time}, Azizi et al. \cite{azizi2023synthetic} incurs over 3.8× training cost to traditional training, while our method increases traditional training by only 30\% yet still outperforms Azizi et al. \cite{azizi2023synthetic}.

\begin{table}[h]
    \vspace{-4mm}
    \centering
    \caption{Total generation and training time on ImageNet dataset.} \label{tab:ab_time}
    \footnotesize
    \setlength{\tabcolsep}{1.5mm}
    \renewcommand{\arraystretch}{1.1}
    \begin{tabular}{c|ccc|c}
    \toprule
    Method & Real only & Azizi [1] & SD Random & Ours \\
    \hline
    Time (GPU hours) & 384 & $>1467$ & 489 & 496 \\
    \bottomrule
    \end{tabular}
    \vspace{-6mm}
\end{table}

\subsection{More Ablation Studies}

\textbf{Performance on different generation numbers.} In Table~\ref{tab:ab_gen_num}, we compare the influence of number of generated images in our method. According to the results, we can see that our method can get improvements with a small amount of generated images such as 1K. While when we keep increasing the number to 1300K (the size of real set is 1280K), the accuracy drops. We state that, there is a trade-off in active learning between prioritizing challenging samples and maintaining the ability for basic samples; overemphasis on challenging samples can potentially lead to a loss of discriminability on those fundamental ones. To address this issue, we increase the diversity of selected samples as the generation number grows by adjusting the threshold probability for sample selection. As the results shown in the last row in the table, after adjusting the selection thresholds of the large generation numbers, their performance increases continuously, as more diverse samples are selected for generation.

\begin{table}[h]
    \vspace{-4mm}
    \centering
    \small
    \setlength{\tabcolsep}{1.2mm}
    \renewcommand{\arraystretch}{1.1}
    \caption{Comparisons of different generation numbers on ResNet-50 and ImageNet. Mis.: we use misclassified samples for generation.}
    \label{tab:ab_gen_num}
    \begin{tabular}{c|ccccccc}
        \toprule
        \#Gen (K) & 0 (real only) & 1 & 10 & 50 & 130 & 650 & 1300 \\
        \midrule
        ACC & 76.39 & 76.52 & 78.24 & 78.51 & 78.65 & 78.11 & 78.03\\
        \midrule
        \multicolumn{8}{c}{Adjustments of threshold}\\
        \midrule
        Threshold & - & mis. & mis. & mis. & mis. & $<0.2$ & $<0.5$\\
        ACC & 76.39 & 76.52 & 78.24 & 78.51 & 78.65 & 78.97 & 79.18\\
        \bottomrule
    \end{tabular}
    \vspace{-4mm}
\end{table}

\textbf{Comparison to focal loss.} Our ActGen proposes an active generation approach to guide the student learning better on the hard samples, while there exist some losses such as Focal Loss~\cite{lin2017focal} that enlarge the loss weights on the hard samples. We now conduct experiments to compare our method with them. As shown in Table~\ref{tab:ab_focal_loss}, we implement Focal Loss ($\gamma=2.0,\alpha=0.25$) and a simple weighted cross-entropy loss that multiples $5\times$ weights on the misclassified samples. The results show that, on ImageNet, both Focal Loss and Weighted CE lead to performance collapse; on CIFAR-10, focal loss gains improvement while still worse than our ActGen. There are some possible explanations to these results: (1) The adaptive weights may cause unstable gradients and the hyper-parameters in these losses are difficult to tune on current sophisticated-designed training strategy. (2) Focusing on some noisy samples (\textit{e.g.}, samples with wrong class annotations) may lead undesired optimization directions and disturb the training. (3) Our ActGen introduce additional diverse images, which are more beneficial for the model to learn the patterns in hard samples.

\begin{table}[h]
    \vspace{-4mm}
    \centering
    \small
    \setlength{\tabcolsep}{1mm}
    \renewcommand{\arraystretch}{1.1}
    \caption{Comparisons with Focal loss and weighted CE.}
    \label{tab:ab_focal_loss}
    \begin{tabular}{c|cc|cc}
        \toprule
        Method & Dataset & Model & ACC & ACC $\Delta$ \\
        \midrule
        Real only & \multirow{4}*{ImageNet} & \multirow{4}*{ResNet-50} & 76.39 & -\\
        Focal Loss & & & 76.03 & -0.36\\
        Weighted CE & & & 75.21 & -1.18\\
        ActGen (ours) & & & \textbf{78.65} & \textbf{+2.26}\\
        \midrule
        Real only & \multirow{4}*{CIFAR-10} & \multirow{4}*{VGG-16} & 95.02$\pm$0.17 & - \\
        Focal Loss & & & 95.27$\pm$0.24 & +0.25\\
        Weighted CE & & & 94.75$\pm$0.39 & -0.27\\
        ActGen (ours) & &  & \textbf{95.53}$\pm$0.37 & \textbf{+0.51}\\
        \bottomrule
    \end{tabular}
\end{table}

\end{document}